\documentclass[11pt]{article}
\usepackage{nodalida2019}
\usepackage{times}
\usepackage{graphicx}
\usepackage{geometry}
\usepackage{url}
\usepackage{booktabs}
\usepackage{latexsym}
\usepackage{enumitem}

\aclfinalcopy 

\title{Neural Cross-Lingual Transfer and Limited Annotated Data\\for Named Entity Recognition in Danish}

\author{Barbara Plank \\
  Department of Computer Science \\
 ITU, IT University of Copenhagen \\
  Denmark \\
  {\tt bplank@itu.dk}
  }

\date{}

\begin{document}
\maketitle
\begin{abstract} 
Named Entity Recognition (NER) has greatly advanced by the introduction of deep neural architectures.
However, the success of these methods depends on large amounts of training data. 
The scarcity of publicly-available human-labeled datasets
has resulted in limited evaluation of existing NER systems, as is the case for Danish.
This paper studies the effectiveness of cross-lingual transfer for Danish, evaluates its complementarity to limited gold data,
and sheds light on performance of Danish NER. 
\end{abstract}

\section{Introduction}
Named entity recognition is a key step for natural language understanding (NLU), and important for information extraction, relation extraction, question answering and even privacy protection.  However, the scarcity of publicly-available human annotated datasets
has resulted in a lack of evaluation for languages beyond a selected set (e.g., those covered in early shared tasks like Dutch, German, English, Spanish),
despite the fact that NER tools exists or recently emerged for other languages. One such  case is Danish, for which NER dates back as early as~\cite{bick2004} and tools exist~\cite{bick2004,derczynski2014,johannessen2005named,polyglot} but lack empirical evaluation.

Contemporarily, there exists  a surge of interest in porting NLU components quickly and cheaply to new languages. This includes cross-lingual transfer methods that exploit resources from existing high-resource languages for zero-shot or few-shot learning. This line of research is blooming, particularly since the advent of neural NER, which holds the state of the art~\cite{yadav2018survey}. However, neither neural tagging nor cross-lingual transfer has been explored for Danish NER, a gap we seek to fill in this paper.
 
\paragraph{Contributions} We present  a) publicly-available evaluation data to encourage research on Danish NER; b) an empirical comparison of two existing NER systems for Danish to a neural model; c) an empirical evaluation of learning an effective NER tagger for Danish via cross-lingual transfer paired with very little labeled data.

\section{Approach}

We investigate the following questions: \textbf{RQ1:} To what extent can we transfer a NER tagger to Danish from existing English resources? \textbf{RQ2:} How does cross-lingual transfer compare to annotating a very small amount of in-language data (zero-shot vs few-shot learning)?  \textbf{RQ3:} How accurate are Danish NER systems?

\subsection{NER annotation}
To answer these questions, we need gold annotated data. Access to existing resources is limited as they are not available online or behind a paywall. Therefore, we annotate NERs  on top of publicly available data.\footnote{\tiny{\url{github.com/UniversalDependencies/UD_Danish-DDT}}}

 In line with a limited budget for annotation
~\cite{garrette2013}, we add an annotation layer for Named Entities to the development and test sets of the Danish section of the Universal Dependencies (UD) treebank~\cite{johannsen2015universal,nivre2016universal}. In particular, all tokens marked as proper nouns in UD were annotated for four named entity types. Annotation were obtained by two annotators with an inter-annotator agreement of 0.9 using Cohen's $\kappa$ (on entities only). To answer RQ2, we further annotate a small portions of the train data: the first 5k and 10k tokens. Table~\ref{fig:ex} shows examples, Table~\ref{ref:stats} provides data statistics. 

The Danish UD treebank (\texttt{Danish-DDT}) is a conversion of the Copenhagen Dependency Treebank (CDT). CDT~\cite{buch-kromann}  consists of 5,512 sentences and 100k tokens, originating from the PAROLE-DK project~\cite{bilgram1998construction}. In contrast to the original CDT and the PAROLE tokenization scheme, starting from Danish UD has the advantage that the language use is closer to everyday language, as it splits tokens which were originally joined (such as `i\_alt').

We follow the CoNLL 2003 annotation guidelines~\cite{sang2003introduction} and annotate proper names  of four types: person (PER), location (LOC), organization (ORG) and miscellaneous (MISC). MISC contains for example names of products, drinks or film titles. 

\subsection{Cross-lingual transfer} 
We train a model on English (a medium and high-resource setup, see details in Section~\ref{sec:exp}) and transfer it to Danish, examining the following setups:
\begin{itemize}[noitemsep]
\item \textbf{Zero-shot}: Direct transfer of the English model via aligned bilingual embeddings. 
\item \textbf{In-language}: Training the neural model on very small amounts of in-language Danish training data only. We test two setups, training on the tiny data alone; or with unsupervised transfer via word embedding initialization (+Poly).
\item \textbf{Few-shot direct transfer}: Training the neural model on English and Danish jointly (on the concatenation of the data), including bilingual embeddings.
\item \textbf{Few-shot fine-tuning}: Training the neural model first on English, and fine-tuning it on Danish. This examines whether fine-tuning is better than training the model from scratch on both.
\end{itemize}

\begin{table}
\resizebox{\columnwidth}{!}{

\begin{tabular}{cccccccccc}
\toprule
B-LOC & O & O & O & O & O & O & O\\
Rom & blev & ikke & bygget & p\r{a} & \`{e}n & dag & .\\
\midrule
O & O & O & B-PER & O & O & B-MISC & I-MISC \\
vinyl & , & som & Elvis & indspillede & i & Sun & Records \\
\bottomrule
\end{tabular}
}
\caption{Example annotations.}
\label{fig:ex}
\end{table}

\begin{table}\centering
\resizebox{.95\columnwidth}{!}{
\begin{tabular}{lrr|rr}
\toprule
                   & \multicolumn{2}{c}{Evaluation} & \multicolumn{2}{c}{Training}\\
                   & \textsc{Dev} & \textsc{Test} & \textsc{Tiny} & \textsc{Small} \\
                   \midrule
Sentences & 564 & 565 & 272 & 604\\
Tokens & 10,332 & 10,023 & 4,669 & 10,069\\
Types & 3,640 & 3,424 &  1,918 & 3,525 \\
TTR & 0.35 & 0.34 & 0.41 & 0.35\\
Sent.w/  NE & 220 & 226& 96& 206 \\
Sent.w/  NE\% & 39\% & 34\% & 35\% & 34\%\\
Entities & 347 & 393 & 153 & 341\\
\bottomrule
\end{tabular}
}
\caption{Overview of the annotated Danish NER data. Around 35\%-39\% of the sentences contain NEs. TTR: type-token ratio.}
\label{ref:stats}
\end{table}

\begin{table*}\centering
\begin{tabular}{l|r|rr|rr|r}
\toprule
                   &       & \multicolumn{2}{c}{neural in-lang.} & \multicolumn{3}{|c}{neural transfer} \\
                   & TnT & plain & +Poly & +\textsc{Medium} src & +\textsc{Large} src & \textsc{FineTune}\\
                   \midrule
zero-shot & --- &  ---  & --- & 58.29 & 61.18 & --- \\
\midrule
\textsc{Tiny} & 37.48 &  36.17 & 56.05 & 67.14 & 67.49 & 62.07\\
\textsc{Small} & 44.30 & 51.90 & 67.18 & \textbf{70.82} & 70.01 & 65.63\\
\bottomrule
\end{tabular}
\caption{F$_1$ score on the development set for low-resource training setups (none, tiny 5k or small 10k labeled Danish sentences). Transfer via multilingual embeddings from \textsc{Medium} (3.2k sentences, 51k tokens) or \textsc{Large} English source data (14k sentences/203k tokens).}
\label{ref:results}
\end{table*}

\section{Experiments}\label{sec:exp}

As source data, we use the English CoNLL 2003 NER dataset~\cite{sang2003introduction} with 
 BIO tagging. 

We study two setups for the source side: a \textsc{Medium} and \textsc{Large} source data setup. 
For \textsc{Large} we use the entire CoNLL 2003 training data as starting point, which contains around 14,000 sentences and 200,000 tokens. 
To emulate a lower-resource setup, we  consider a \textsc{Medium} setup, for which we employ the development data from CoNLL 2003 as training data (3,250 sentences and 51,000 tokens).  The CoNLL data contains a high density of entities (79-80\% of the sentences) but is lexically less rich (TTR of 0.11-0.19), compared to our Danish annotated data (Table~\ref{ref:stats}), which is orders of magnitudes smaller, lexical richer but less dense on entities. The difference in entity richness might stem from the fact that the CoNLL data was explicitly collected for NER, while the UD was collected in an effort to create a syntactically annotated treebank.

\paragraph{Model and Evaluation} We train a bilstm-CRF similar to~\cite{xie2018neural,johnson-etal-2019-cross,plank-etal-2016-multilingual}. As pre-trained word embeddings we use the Polyglot embeddings~\cite{polyglot}. The word embeddings dimensionality is 64. The remaining hyperparameters were determined on the English CoNLL data. The word LSTM size was set to 50. Character embeddings are 50-dimensional. The character LSTM is 50 dimensions. Dropout was set to 0.25. We use Stochastic Gradient Descent with a learning rate of 0.1 and early stopping. We use the evaluation script from the CoNLL shared task and report mean F1 score over three runs.

\paragraph{Cross-lingual mapping} We map the existing Danish Polyglot embeddings 
to the English embedding space by using an unsupervised alignment
method which does not require parallel data. In particular, we use  
character-identical words as seeds for the Procrustes rotation method introduced in \texttt{MUSE}~\cite{conneau2017word}.

\section{Results}\label{sec:results}

Table~\ref{ref:results} presents the main results. There are several take-aways.

\paragraph{Cross-lingual transfer} is powerful (RQ1). Zero-shot learning reaches an F1 score of 58\% in the \textsc{Medium} setup, which outperforms training the neural tagger on very limited gold data (plain). Neural NER is better than traditional HMM-based tagging (TnT)~\cite{brants2000tnt} and greatly improves by unsupervised word embedding initialization (+Poly). It is noteworthy that zero-shot transfer benefits only to a limiting degree from more source data  (F1 increases by almost 3\% when training on all English CoNLL data). 

To compare cross-lingual transfer to limited gold data (RQ2), we observe that training the neural system on the small amount of data together with Polyglot embeddings is close to the tiny-shot transfer setup.  Few-shot learning greatly improves over zero-shot learning. The most beneficial way is to \textit{add} the target data to the source, in comparison to fine-tuning. This shows that 
access to a tiny or small amount of training data is effective.  Adding gold data with cross-lingual transfer is the best setup. In both \textsc{Medium} and \textsc{Large} setups are further gains obtained by adding \textsc{Tiny} or \textsc{Small} amounts of Danish gold data. Interestingly, a) fine-tuning is less effective; b) it is better to transfer from a medium-sized setup than from the entire CoNLL source data.  

\paragraph{Existing systems} (RQ3) are evaluated and results are shown in Table~\ref{ref:existing}. Polyglot~\cite{polyglot} overall performs better than DKIE~\cite{derczynski2014}.\footnote{We use Polyglot version 16.07.04 and the DKIE model from July 23, 2019 commit {6204e1f7245347fe3729ee5d0960380d64ae64e2}.}  The best system is our cross-lingual transfer NER from \textsc{Medium} source data paired with \textsc{Small} amounts of gold data. Per-entity evaluation shows that the neural Bilstm tagger outperforms Polyglot except for Location, which is consistent across evaluation sets. Overall we find that very little data paired with dense representations yields an effective NER quickly.

\begin{table}\centering
\resizebox{\columnwidth}{!}{

\begin{tabular}{lrrrrr}
\toprule
\textsc{Dev}    & All & PER & LOC & ORG & MISC \\
                 \midrule

Majority & 44.4 & 61.8 & 0.0 & 0.0 & ---\\
DKIE  & 58.9 & 68.9 & 63.6 & 23.3  &--- \\
Polyglot &64.5 & 73.7 & \textbf{73.4} & 36.8& ---\\
Bilstm & \textbf{70.8} & \textbf{83.3} & 71.8 & \textbf{60.0}  & 23.9\\

\midrule
\textsc{Test}    & All & PER & LOC & ORG & MISC \\
Polyglot & 61.6 & 78.4 & \textbf{69.7} & 24.7& ---\\
Bilstm & \textbf{66.0} & \textbf{86.6} & 63.6& \textbf{42.5} & 24.8\\

\bottomrule
\end{tabular}
}
\caption{F$_1$ score for Danish NER.} 
\label{ref:existing}
\end{table}

\section{Related Work}

Named Entity Recognition has a long history in NLP research. While interest in NER originally arose mostly from a question answering perspective, it developed into an independent task through the pioneering shared task organized by the Message Understanding Conference (MUC)~\cite{grishman-sundheim-1996-muc,grishman-1998-research}.  Since then, many shared task for NER have been organized, including CoNLL~\cite{sang2003introduction} for newswire and WNUT for social media data~\cite{baldwin2015shared}. 
While Danish NER tools exists~\cite{bick2004,derczynski2014,johannessen2005named,polyglot}, there was a lack of reporting F1 scores~\cite{lacunae}. Supersense tagging, a task close to NER, has received some attention~\cite{martinez-alonso-etal-2015-supersense}.

The range of methods that have been proposed for NER is broad. Early methods focused on hand-crated rule-based methods with lexicons and orthographic features. They were followed by feature-engineering rich statistical approaches~\cite{nadeau2007survey}. Since the advent of deep learning and the seminal work by~\newcite{collobert2011natural}, 
state-of-the-art NER systems typically rely on feature-inferring encoder-decoder models that extract dense embeddings from word and subword embeddings, including affixes~\cite{yadav2018survey}, often outperforming neural architectures that include lexicon information such as  gazetteers. 

Recently, there has been a surge of interest in cross-lingual transfer of NER models~\cite{xie2018neural}. This includes work on transfer between distant languages~\cite{rahimi2019multilingual}  and work on projecting from multiple source languages~\cite{johnson-etal-2019-cross}. 

\section{Conclusions}

We contribute to the transfer learning literature by providing a first study on the effectiveness of exploiting English NER data to boost Danish NER performance.\footnote{\url{github.com/bplank/danish_ner_transfer}}
We presented a publicly-available evaluation dataset and compare our neural cross-lingual Danish NER tagger to existing systems.  Our experiments show that a very small amount of in-language NER data pushes cross-lingual transfer, resulting in an effective Danish NER system.

\section*{Acknowledgements}

We kindly acknowledge the support of NVIDIA
Corporation for the donation of the GPUs used in this research and
Amazon Corporation for an Amazon Research Award.

\bibliographystyle{acl_natbib}
\bibliography{nodalida2019}

\end{document}